\documentclass[nonacm]{acmart}
\AtBeginDocument{\providecommand\BibTeX{{\normalfont B\kern-0.5em{\scshape i\kern-0.25em b}\kern-0.8em\TeX}}}

\usepackage{graphicx}
\usepackage{multirow}
\usepackage{threeparttable}
\usepackage{relsize}
\usepackage{caption}
\usepackage{subcaption}

\usepackage{xcolor}

\setcopyright{none}

\begin{document}

\title{An RNN-Survival Model to Decide Email Send Times}

\author{Harvineet Singh}
\authornote{Work done during an internship at Adobe Research.}
\email{hs3673@nyu.edu}
\affiliation{\institution{New York University}
}

\author{Moumita Sinha}
\email{mousinha@adobe.com}
\affiliation{\institution{Adobe Research}
}

\author{Atanu R. Sinha}
\email{atr@adobe.com}
\affiliation{\institution{Adobe Research}
}

\author{Sahil Garg}
\authornotemark[1]
\email{sahil.garg@nutanix.com}
\affiliation{\institution{Nutanix}
}

\author{Neha Banerjee}
\authornotemark[1]
\affiliation{\institution{Adobe Systems}
}

\begin{abstract}
Email communications are ubiquitous. 
Firms control send times of emails and thereby the instants at which emails reach recipients (it is assumed email is received instantaneously from the send time). However, they do not control the duration it takes for recipients to open emails, labeled as time-to-open. Importantly, among emails that are opened, most occur within a short window from their send times. We posit that emails are likely to be opened sooner when send times are convenient for recipients, while for other send times, emails can get ignored. Thus, to compute appropriate send times it is important to predict times-to-open accurately. We propose a recurrent neural network (RNN) in a survival model framework to predict times-to-open, for each recipient. Using that we compute appropriate send times. We experiment on a data set of emails sent to a  million customers over five months. 
The sequence of emails received by a person from a sender is a result of interactions with past emails from the sender, and hence contain useful signal that inform our model. This sequential dependence affords our proposed RNN-Survival (RNN-S) approach to outperform survival analysis approaches in predicting times-to-open. We show that best times to send emails can be computed accurately from predicted times-to-open. This approach allows a firm to tune send times of emails, which is in its control, to favorably influence open rates and engagement.

\end{abstract}

\maketitle

\section{Introduction}
\label{secIntroduction}

Ubiquity of email for communication makes recipients feel overwhelmed by the number of messages received ~\cite{gupta2016email}. Consequently, non-work emails such as marketing messages may get opened only when it is convenient for recipients. While a marketer cannot control which times are convenient for recipients to open messages, it controls send times of messages. A marketer can tune send times to make them convenient for recipients to open emails. We show how to determine appropriate send times and do so at the individual level. Throughout the paper `send time' means both the send time and the time an email arrives in an inbox, under the premise that emails arrive instantaneously.
Research in behavior modeling of email users \cite{yang2017} has been enabled by the Enron corpus \cite{klimt2004enron} and Avocado research email collection \cite{oard2015avocado}. 
While researchers recognize interaction behaviors, send times of emails have not been examined, although these impact responses and interaction behaviors of recipients.

Consider permission emails where customers sign up to receive marketing emails from a firm. Open rates remain typically low -- ranging from 15\% to 19\% in e-commerce, beauty and personal care, 
and in the range of 20\% to 28\% in hobbies, 
health and fitness industries \cite{smartinsights}. These aggregate measures mask the variability that individuals show in the duration to open emails from their send times. If data show that variability in open rate is systematically associated with variability in duration to open, it behooves recognizing the latter in examining interaction behaviors, yet has largely been ignored. First, we model the duration it takes to open emails, labeled as \textit{time-to-open}. For higher open rates, emails ought to arrive in inboxes at times that are favorable for opening emails. As a second step, delivery of emails at favorable time periods is achieved by selecting send times.
In sum, we determine appropriate send times by learning to predict times-to-open.

\begin{figure}[!htbp]
  \centering
  \includegraphics[width = 0.75\columnwidth, keepaspectratio]{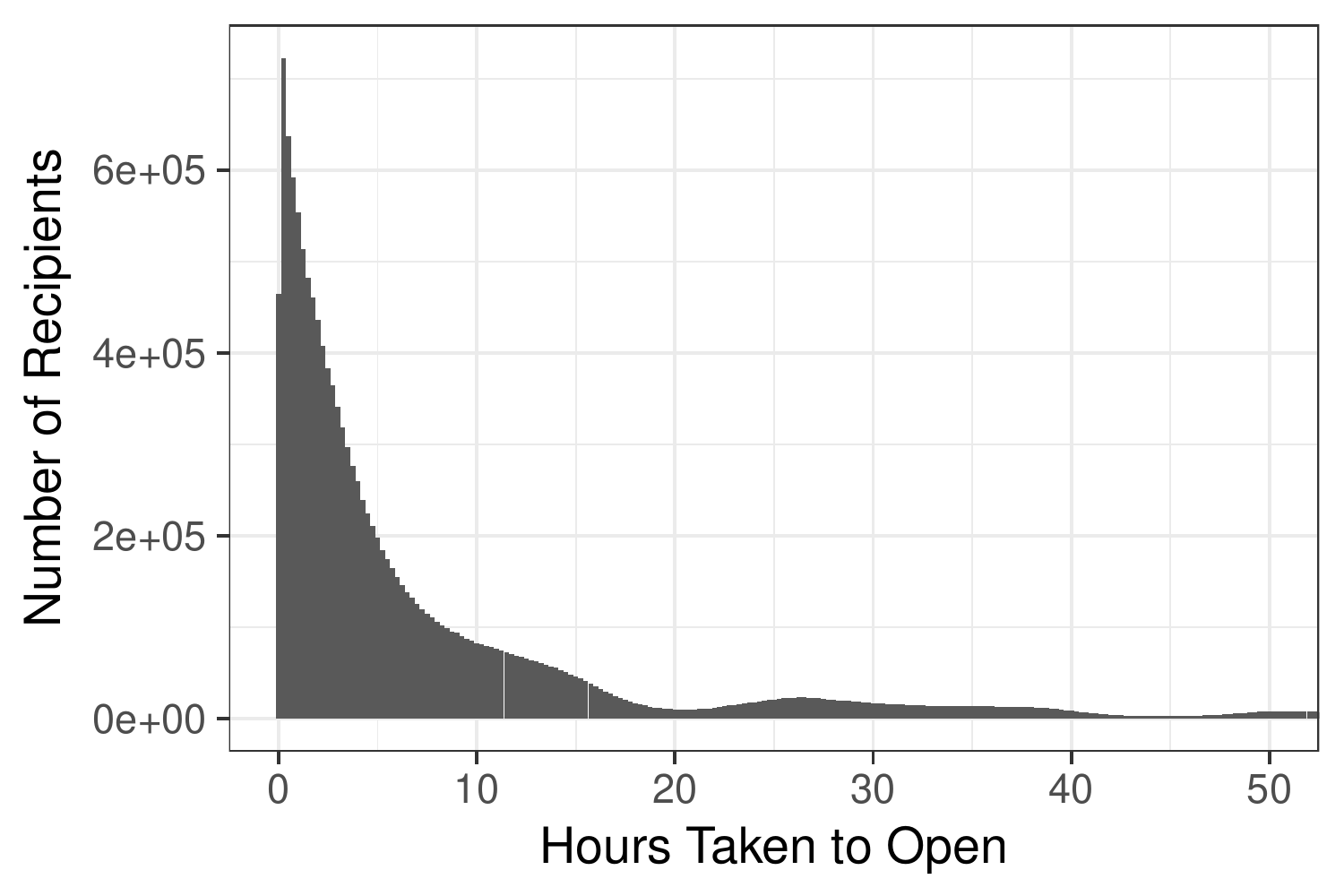}
  \caption{Time taken to open messages, after they were sent.}
  \label{fig:time}
\end{figure}

We first examine the distribution of times-to-open for our email data set (Figure~\ref{fig:time}), which is based on send times at different hours of day and different days of week. Overall, it shows that if emails are not opened within a relatively short duration window the likelihood of being opened diminishes considerably. Hence, it becomes crucial to identify send times that are convenient for the recipient to open emails. However, determining duration of time poses a challenging problem because in the observation period most emails are not opened; yet, additional emails could be opened later. This censoring of time-to-open, calls for survival models. Traditional survival models are not helpful since they fail to account for the sequential dependence of email interactions a recipient shows and the consequent impact on time-to-open for future emails. E.g., if a user reads a few emails successively it likely affects whether and when she reads the email arriving next. This could be due to time constraint. An additional challenge is that the time to open varies by individuals and any model ought to capture heterogeneity among users. This in turn requires any implementable model to account for a degree of personalization. From the firm's perspective, \textit{time-to-open} provides an estimate of interest-level of an individual for a firm's campaigns. A consumer who opens marketing emails more frequently is more engaged with brands and hence lower time-to-open indicates higher degree of interest. By contrast, open rates computed regardless of duration in which those are opened, fail to differentiate between persons with the same open rate but who do so in a short duration (more engaged and more interested in brand's messages) versus those who use a long duration (less engaged and less interested in brand's messages). Thus, time-to-open along with open rate yield richer behavioral information than aggregate metric such as open rate, which is traditionally used for measuring interest-levels.

\textbf{Contributions}. The main contributions of our paper are: (1) modeling time to open an email, in a survival analysis framework, which can identify a personalized time to send an email to an individual, with improved predictive performance for opening the email, (2) modeling the inherent, sequentially time dependent nature of emails and incorporating RNN to address this dependence within survival models, yielding improved performance versus baselines, and (3) de-seasonalizing the data for incidence of emails, through a new technique for binning, which is grounded in literature and its use in our RNN-S model provides seasonality adjusted results.

\section{Survival Analysis}
\label{secSurvivalAnalysis}

Survival models are widely used in many disciplines from biomedical science \cite{collett2015modelling} to education \cite{ameri2016survival} to marketing science \cite{manchanda2006effect}. Their usefulness lies in handling two characteristics: one, the event of interest is not observed for some individuals within the data observation window; two, time-varying features are important for inclusion in the model. In this paper, our interest lies in time to open emails. The random variable time-to-open is observed for emails that are opened in the observation window but is right-censored for emails that are not opened in the observation window. If certain features of emails vary with time over the observation window and affect time-to-open, the model needs to reflect that. Regression models are inadequate at handling either characteristic and cannot be a baseline. Also note that survival models are not akin to probabilistic models for binary outcomes, such as a logistic regression (LR), since in LR the intrinsic probability that an event occurs is constant over time. Whereas in a survival duration model this probability is allowed to change with time. Moreover, LR does not take into account the censored information and does not predict on a continuous scale time-to-event. Thus, we cannot use LR model as a baseline. 
To account for whether an email is opened at all we consider only the first instance of opening an email, not the repeat instances. Let $T$ represent time-to-open for the first instance, with probability density function $f(t)$ and distribution function $F(t)$ at time $t$. The survivor function $S(t)$ represents the probability that an individual opens email after time $t$ and is given by $S(t)=P(T\geq t)=1-F(t)=\int\limits_{t}^{\infty} f(v) dv$. To compute the probability of an email being opened for the first time by a recipient, we use the hazard function $h(t)$, which measures the instantaneous probability that an individual opens an email in time $t + \delta t$, conditional on not having opened till time $t$. The hazard function is given by $h(t)=\lim_{\delta t\to 0} \frac{P(t \leq T <t+ \delta t| T \geq t)}{\delta t}$. For details of survival and hazard functions see  \cite{sinha2018modeling, liu2010understanding, collett2015modelling}.

\subsection{Cox Proportional Hazard Regression}
\label{coxsection}
Following \cite{ameri2016survival}, \cite{manchanda2006effect},
we use Cox Proportional Hazard regression (CPH) to incorporate censoring information in predicting time to open an email. 
At time $t$, the hazard function for the  $i^{\text{th}}$ individual, dependent on feature vector $X_i$, is given by $h_i(t|X_i)=h_0(t) \times \phi(X_i)$. Here $h_0(t)$ is the baseline hazard at $t$, same for all individuals, and $\phi(X_i)$ is the hazard ratio \cite{wang2017machine}, defined as the hazard function of the $i^{\text{th}}$ individual compared to that of another individual whose values for all the features considered in the model are zero. Note $\phi(X_i)$ is independent of time and depends only on features.

Moreover, our data suggest that individuals who are likely to open marketing emails at all form a minority class. To account for this heterogeneity in intrinsic propensities across classes of individuals who are likely to open and who are not likely to open, mixture models are used \cite{ farewell1982use}. The survival function is modified as a weighted ($w_i$) sum of survival functions for the group which is likely to open (\textit{SG}) and for the group unlikely to open such emails and is given by $S_i(t|X_i)=w_iS(t_i|SG=1,X_i) + (1 - w_i)$ \cite{ farewell1982use}.

\begin{figure*}[t]
  \centering
  \begin{subfigure}{0.25\textwidth}
        \centering
        \includegraphics[width = 1.0\textwidth, keepaspectratio]{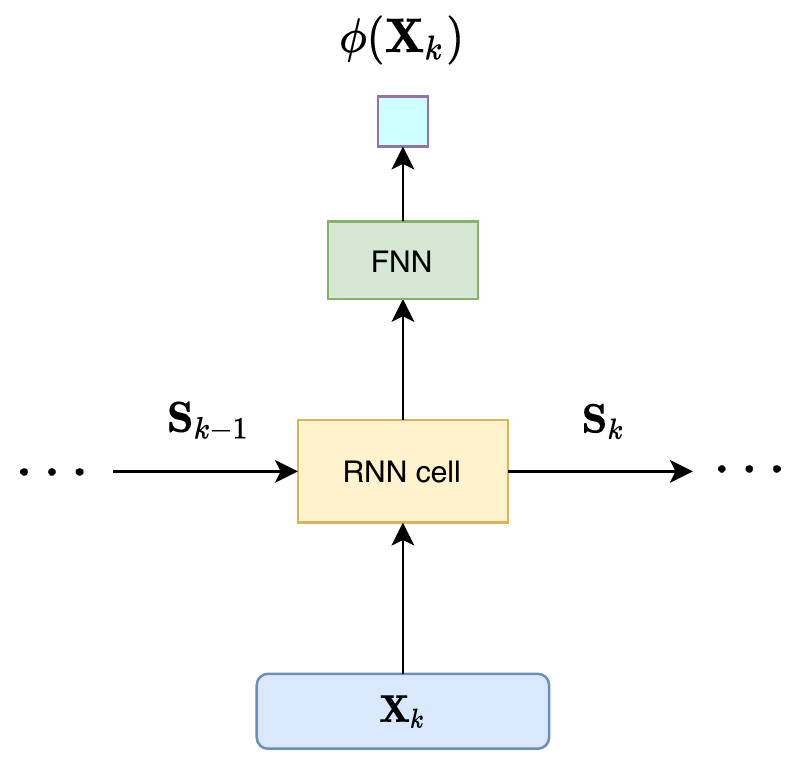}
        \caption{}
        \label{fig:rnn}
    \end{subfigure}\begin{subfigure}{0.75\textwidth}
        \centering
        \includegraphics[width = 0.9\textwidth,keepaspectratio]{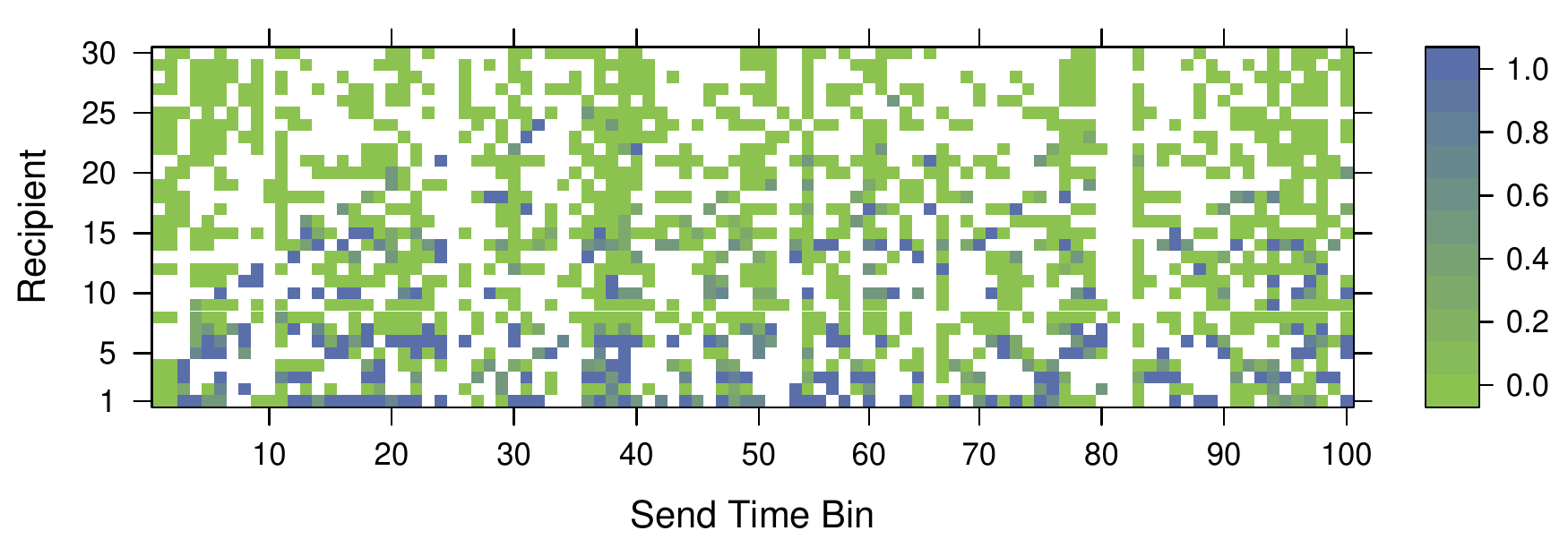}
        \caption{}
        \label{fig:or}
    \end{subfigure}
    \caption{(a) Recurrent neural network to predict hazard rate. (b) Open rates for different send time bins by recipients. Plot shows variability in open rates across bins and recipients.}
\end{figure*}

\subsection{RNN with Survival Analysis}
\label{secrnn}

In CPH regression, past interaction information is incorporated as a feature vector assuming that interactions are independent of one another. This ignores that a stream of emails from a firm is an outcome of the history of interactions of the individual with past emails from this firm. Our recognition of dependence in the sequential interaction information is a significant departure from CPH. Recurrent Neural Network (RNN) models the dependence by taking as input the hidden state vector obtained from past elements along with the input vector for the current element to determine outputs, which are usually modeled as a function of hidden state vector $s_t$. Many different variations of the non-linear transformation function exist such as long short-term memory (LSTM) \cite{hochreiter1997long} and gated recurrent unit (GRU) \cite{chung2014empirical}.

\textbf{Proposed model}: To incorporate RNN in survival analysis framework, the model architecture considered, is shown in Figure \ref{fig:rnn}. The input sequence is first passed through an RNN cell one message at a time, per individual, followed by a feedforward neural network (FNN). The result of the FNN is passed through the exponential activation function so that the final prediction for hazard rate is always positive. For each message, RNN cell takes as input the state vector of the cell at the last message for the same individual, thus incorporating the sequentially time dependent information for each individual. Consider that the $i^\text{th}$ individual receives $k_i$ messages during the observation window.
For the  $i^\text{th}$  individual and $j^\text{th}$ message, the feature set, outcome and censoring information are denoted by $(X_{ij}, Y_{ij}, I_{ij})$. The output layer of the network predicts the hazard ratio, $\phi(X_{ij})$, for $j^\text{th}$ message and $i^\text{th}$ individual. Thus, we express the training dataset as $\mathcal{D} := \{(\phi(X_{ij}), Y_{ij}, I_{ij}) | i=1 \dots n, j=1 \dots k_i\}$. 

RNN conditions its output on information from the history of past messages, and here it represents the conditional distribution of the hazard rates given past messages, for each message and for each individual. Thus the likelihood of responses for an individual can be computed directly using RNN outputs for each of the messages for the individual. Assuming independence of responses across individuals, the overall likelihood becomes the product of likelihood for each individual. 

The negative log partial likelihood with Efron's approximation \cite{efron1977efficiency} is used as the loss function. Efron's approximation modifies the partial likelihood function
to account for data points with tied times to open. Consider the data has $N$ observed unique times to open with the $l^\text{th}$ open time denoted by $t_l$ such that $l= 1 \dots N$ and $N\leq{\sum \limits_{i}k_i}$. The partial likelihood function is then written as:

\begin{equation}
\label{eq:loglik}
    L(\beta)= \mathlarger{\prod \limits_{l=1}^{N}} \frac{\mathlarger{\prod \limits_{h \in D_l} \phi(X_h; \beta)}}{\mathlarger{\prod\limits_{m=1}^{d_l}} [\mathlarger{\sum \limits_{h \in R_l} \phi(X_h; \beta) - \frac{m-1}{d_l} }\mathlarger{\sum \limits_{h \in D_l}} \phi(X_h; \beta)]}
\end{equation}

where $R_l$ is the set of observations which are yet to open the message at time $t_l$, $D_l$ is the set of observations which are tied at time $t_l$, and $d_l$ is the total number of observations which are tied at that time. Here $\phi(X_h; \beta)$ denotes the predicted hazard rate for a message of an individual, jointly indexed as $h$, given the parameters of the network $\beta$.

\section{Related Work}
\label{secRelatedWork}

Survival analysis has found applications in website and online technology, such as in determining dwell time on a website \cite{liu2010understanding}, emails \cite{sinha2018modeling}, in predicting listening behaviour on music websites \cite{jing2017neural}, and in estimating reciprocal-link creation time in online social networks \cite{dave2017fast}. Using medical data recent work has studied applications of Gaussian processes \cite{alaa2017deep} and latent variable models \cite{ranganath2016deep} in survival analysis.
Past work using email interaction data includes prediction of response time \cite{sinha2018modeling, yang2017enterprise} and real-time evaluation of campaigns \cite{bonfrer2009real}. From a methodological perspective, survival analysis has found applications in marketing in predicting customer life-time value \cite{lu2003modeling} and customer churn \cite{jamal2006improving}. 

Recent efforts using deep learning have shown improvements in modeling survival data \cite{katzman2018deepsurv,ren2019deep,lee2018deephit,luck2017deep}.
In modeling time-to-open an email, \cite{sinha2018modeling} aggregate individual's interactions with and responses to previous emails, while ignoring the sequential time dependence among emails.
In scenarios with a history of email message interactions between individuals and the firm, previous work does not recognize that these interactions are sequentially time dependent, which we do through RNN within survival model. 
Jing et al. \cite{jing2017neural} propose the use of survival analysis with RNN for an online music service. In-session activity of users and the time they spend on the music service are used for modeling return times, which makes our setting different as corresponding information are not available from email activity. Similarly, Ren et al. \cite{ren2019deep} showed improvements from sequential modeling for three domains---real-time bidding, music streaming, and health. WTTE-RNN \cite{martinsson2016wtte} applies RNNs for predicting the time taken for a customer to churn, an objective function which is different from ours.
 \section{Problem Definition and Data Descriptions}
\label{secProblemDef}

We first provide model-free evidence that send time has a bearing on when an email is opened and then describe the data used.

\textbf{Encoding of Send time bins}. The data comprise of marketing email messages sent to customers who signed up to receive emails. The send times are controlled by the marketer and suffer from seasonality over hours of day and days of week. Fewer messages are sent during late night to early morning hours, and more messages are sent on weekdays relative to weekend-days. It is useful to deseasonalize the data of messages sent over time. Following the concept used by \cite{bonfrer2009real}, we consider one week as the horizon and for each week of data, 168 hours are split into $100$ time-bins such that each bin has equal number of messages sent. The choice of $100$ bins is judgmental and can be controlled by the marketer. Labeled \textit{virtual} send time bins, their use is justified following \cite{bonfrer2009real} as ``Virtual time involves adjusting the speed of time'' through the week. 
Under the premise that emails sent at a convenient time for the recipient are likely to be opened quicker than emails sent at an inconvenient time, the concept of virtual time stretches a duration of actual time that is convenient for the firm and the recipients and compresses an equal duration of actual time when it is not convenient.
Although we have equal number of messages in each bin, the bin duration varies from $13$ hours to $10$ minutes. Notably, this approach to deseasonalization makes it possible to apply our survival model without further complexity in model specification.

\textbf{Relation between Open Rates and Send Times}. As model free evidence, we examine at the individual level, whether send time is important for emails to be opened. In a natural experiment in the data we randomly sample $30$ individuals who received a minimum number of $10$ emails in the observation period of $5$ months. The send times belong to one of the $100$ bins, shown on the x-axis of Figure ~\ref{fig:or}. On the y-axis, the actual open rates of emails sent in each bin are plotted for the $30$ individuals. For a send time bin, the color represents the proportion of received emails that are opened (pure-blue for all emails being opened and pure-green for none of the emails being opened).
The white spaces denote time bins with no emails received. Observing the colors row-wise for each individual,
we find that while some rows show largely green colors (Recipients 25-30), other rows show mix of green and blue (Recipients 10-15), yet others tend to be more blue (Recipient 1). There is a systematic pattern to when individuals open marketing emails. There is also considerable variation across individuals in their propensity to open emails across time bins. Pure green colors suggest that some individuals do not tend to open emails at all, whereas combination of blue and green colors indicate that other individuals open emails sent during specific time bins, but ignore emails sent during other time bins. A few individuals, such as Recipient 1, seem to open most emails across all bins. Armed with this model free evidence we posit that targeting appropriate send times can benefit marketers in increasing open rates.

\textbf{Dataset}. The data contain $1.2$ million customers, interacting with email campaigns sent by an e-retail company during a period of $5$ months (December 2013 to April 2014). For business reasons, which we do not know, the data show variations in number of emails sent to different customers.
We consider only those customers who received at least $10$ messages during the $5$ month period, or received at least $1$ message every $30$ days. This ensures that our predictive model is based on customers who receive regular emails from the retailer.
An email message sent to very few recipients tends to be a one-off message (e.g., birthday greetings) and may have a different response pattern than other email marketing messages. For this reason, emails sent in bulk to less than $400$ recipients are not considered. Additionally, we remove those data where customers have opened a single email more than $10$ times as these likely resulted from non-human agents. The final data comprise of $0.8$ million recipients, $74.5$ million emails, $2,600$ campaigns, 20.8\% open rate, and 2.3\% click rate given opened.

For each email and each customer, the data contain the send time; open time, if opened; and click time on any URL in the email.
Moreover, the data have frequency of product purchase by a customer, purchase date, and whether purchase is online or offline. No personally identifiable information of customers, such as home address, email id, birthday etc. are used in the models explored.

From the data two different representative samples are drawn to perform two separate experiments. The first experiment (detailed in Section \ref{sec:perf_time_to_open}) compares the proposed RNN-Survival model with baseline models to show efficacy of different survival functions. 
For this, a sample of $50,000$ customers is drawn with a train : test split of $80\%$ : $20\%$.  
To incorporate the sequentially time dependent nature of emails, we consider up to $16$ most recent email messages during the $5$ month period. 
More than $70\%$ recipients received at least $16$ messages, which are used for modeling the survival function.

In the second experiment (detailed in Section \ref{sto_section}), results from the predictive model for time-to-open are used to predict best send times. 
For this, we draw a random sample of $30,000$ customers, with each customer receiving at least $50$ messages. This sample is \textit{different} from the data set used to train and test the RNN-Survival model. The first $16$ messages are used to score the predictive model for estimating the send times. Among the remaining messages, percentage of messages opened in each predicted send time bin is calculated to test the model's performance.

 \section{Baseline Models}
\label{models}

As baselines, we use several pure survival models along with features extracted from interactions, but without recognizing any sequential dependence of emails. As mentioned in Section ~\ref{secSurvivalAnalysis}, survival models are suitable baselines because they satisfy three properties of the phenomenon we examine: one, provide predictions on a continuous scale for time to open; two, allow probability of opening an email to change over time; and three, handle censored data. Regression and logistic regression models do not satisfy these properties, making them unsuitable as baselines. 
We consider four baseline models, where past interaction information is incorporated as features for each message, that are mapped to a single feature vector using element-wise averaging. 
Note that averaging includes past open rate (average of open indicators) as a feature in the baselines.

\textbf{Parametric Weibull}.
The survival function follows a Weibull distribution and is given by $S(t) = exp(-\lambda t^\gamma)$, where $\lambda$ is a scale parameter and $\gamma$ is a shape parameter and the distribution is positively skewed. The corresponding hazard function is $h(t) = \lambda\gamma t^{\gamma-1}$. This model assumes that the hazard function increases or decreases (based on $\gamma$) monotonically with increasing survival time.

\textbf{Cox-Proportional Hazards Models}.
Three other baseline models are derived from Cox proportional (CPH) models described in Section \ref{coxsection}. In CPH models, the actual form of the baseline hazard function is not estimated, which is the proportionality constant across individuals at any particular time. The CPH-L and the CPH-G are two variants of CPH, with $\phi(X)$ having a linear and a non-linear relation with features, respectively. The CPH-MM model assumes a mixture of different types of recipient, as explained in Section ~\ref{coxsection}.

\section{Experiments and Results}
\label{secExperiments}

\begin{table*}[t]
\centering
\caption{Comparison of the Models under C-Index across Censoring Windows and Number of Past Emails}
\label{tblSummary}
\resizebox{\linewidth}{!}{\begin{threeparttable}
\begin{tabular}{l|lllll|lllll|lllll}
\toprule
               & \multicolumn{5}{l|}{\textbf{Censoring Window = 3 hours}} 
               & \multicolumn{5}{l|}{\textbf{Censoring Window = 6 hours}}  
               & \multicolumn{5}{l}{\textbf{Censoring Window = 12 hours}}  \\
\textbf{Model} & \textbf{W} &  \textbf{CPH-L} & \textbf{CPH-G} & \textbf{CPH-MM} & \textbf{RNN-S} 
               & \textbf{W} &  \textbf{CPH-L} & \textbf{CPH-G} & \textbf{CPH-MM} & \textbf{RNN-S} 
               & \textbf{W} &  \textbf{CPH-L} & \textbf{CPH-G} & \textbf{CPH-MM} & \textbf{RNN-S} \\
\midrule
\textbf{Number of past Emails = 16}   & 0.8371  & 0.8382 & 0.8429 & 0.8373 & \textbf{0.8438} 
               & 0.8399  & 0.8403 & 0.8468 & 0.8398 & \textbf{0.8539}
               & 0.8426  & 0.8433 & 0.8496 & 0.8432 & \textbf{0.8616} \\
\midrule
\textbf{Number of past Emails = 8}  & 0.8372 &  0.8293 & 0.8392 & 0.8289 & \textbf{0.8409} 
                  & 0.8388 & 0.8291 & 0.8408 & 0.8294 & \textbf{0.847} 
                  & 0.8399 &  0.8303 & 0.8429 & 0.8316 & \textbf{0.8503} \\
\midrule
\textbf{Number of past Emails = 4}  & 0.8022  & 0.8020 & 0.8084 & 0.8020 & \textbf{0.8174} 
                  & 0.8091  & 0.8065 & \textbf{0.8158} & 0.8071 & 0.8108  
                  & 0.8137  & 0.8097 & 0.8184 & 0.8106 & \textbf{0.8271} \\
\bottomrule
\end{tabular}
\end{threeparttable}
}
\end{table*}

In this section, we evaluate the RNN-S model against the baselines,  
and then show that predicted times-to-open can be utilized to compute best send times, which increase open rates of emails.

\textbf{Feature Extraction}. The goal is to capture response to prior email sent, most recent online and offline purchase behaviors, and send time of email. The features we extract are: whether last email opened, whether last email clicked, whether last email mass-mailed (sent to greater than $75$\% of recipients), days since last email sent, days since last purchase, whether Web purchase made since last email, whether offline purchase made since last email, whether direct mails (posts) received since last email,
send time bin of current email ($100$-dimensional one-hot encoded vector).

\textbf{Training details}. RNN-S is trained to minimize negative log partial likelihood using stochastic gradient descent with Adam optimizer \cite{kingma2014adam}. The learning rate is initialized at $0.001$ and batch size is fixed at $32$. To consider varying number of emails sent to recipients, we fix the sequence length for the training examples. Sequences are truncated, if they are longer than the fixed sequence length or padded with zero values if the length of sequence is shorter. In experiments, we show the effect of varying sequence length from $4$ to $16$. We use Long Short-Term Memory (LSTM) variant of the RNN cell with output size of $32$. A single fully connected layer with exponential activation function is used for the FNN shown in Figure \ref{fig:rnn}.

\textbf{Evaluation Metric}.
To compare performance of survival models, we use Concordance Index (C-Index) \cite{harrell1982evaluating, wang2017machine}, a generalization of Area Under the ROC curve to data with censored observations. For C-Index, all pairs of observations are used, excluding those, where both are censored observations. The pairs used are called admissible pairs. The C-Index is the ratio of the number of admissible pairs where the observed order of time to event matches with that of the predicted order, and the total number of admissible pairs. 
A C-Index of 1 means a perfect match for all the admissible pairs and that of 0.5 means a random match.

\begin{figure}[!htbp]
  \centering
  \includegraphics[width = 0.9\columnwidth, keepaspectratio]{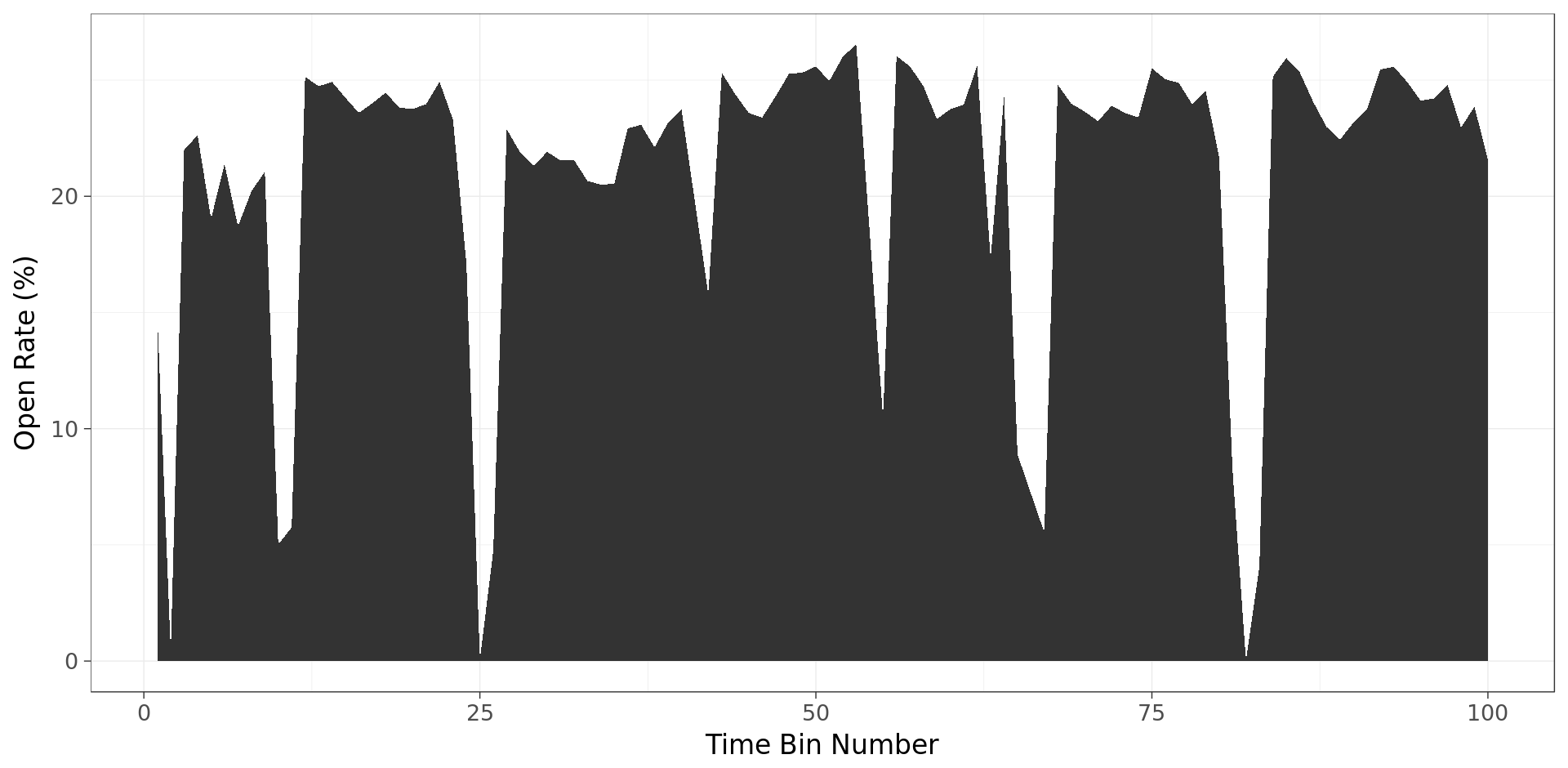}
  \caption{Open Rates by Bins}
  \label{fig:sto}
\end{figure}

We use ranked bins of send times to evaluate the prediction of best times to send messages. Figure \ref{fig:sto} shows the open rates of emails in each bin. By construction, since all time bins have equal number of emails sent, the open rates are expected to be equal across bins, provided times of sending emails are not important. However, Figure \ref{fig:sto} exhibits that open rates are unequal across bins. Hence, to identify the appropriate send times we rank-order bins for each individual based on predicted time-to-open emails, when sent during each bin. Then, we use 
out-of-sample testing data, to compare open rates of emails sent during predicted top ranked time bins of an individual versus those sent during other time bins.

\subsection{Model Performance for Time to Open}
\label{sec:perf_time_to_open}

The performance of RNN-S is evaluated against baselines across two dimensions. First, we consider $3$ different durations for censoring window, namely $3$, $6$  and $12$ hours. Varying window durations helps determine the duration of time, after email is sent, that is needed to reliably predict time to open. Second, we vary the length of sequence of past emails, namely $4$, $8$ and $16$ to uncover the relation between length of historical information and prediction accuracy. In turn, this can establish an empirical lower bound on the length that is necessary to provide desired accuracy of prediction. Table \ref{tblSummary} shows C-Index across different variations, for different models. 

\paragraph{\textbf{Comparison across Censoring Windows}}.
From Table \ref{tblSummary} it is observed that, as the duration of censoring window is increased (see columns), the accuracy increases, with the highest values of C-Index observed for $12$ hours. 
Notably, two insights emerge. One, given a length of sequence (see rows), for every window duration, RNN-S outperforms all four baselines W, CPH-L, CPH-G and CPH-MM in values of C-Indices (except for 1 case). Two, as window duration is increased from $3$ to $12$ hours, the increase in C-Index is more marked for RNN-S, relative to four baselines. In summary, by recognizing sequential time dependence of emails, RNN-S performs better in predicting time-to-open compared to the four baselines, which come from prior art but ignore this time dependence. 
The longest duration we experimented with is 12 hours and could be closer to observation windows of interest to firms in determining times to open, if they monitor email performance daily. 

\paragraph{\textbf{Comparison across Lengths of Sequence}}. Table \ref{tblSummary} shows that the monotonically increasing relation between C-Index and length of sequence holds across models. However, the increase in C-Index between using $4$ past emails and $8$ past emails is higher than that between $8$ and $16$ past emails. Thus, when the length of sequence of past emails is shorter, the RNN-S approach, which helps capture intra-individual variability, provides higher lift in accuracy than other baseline survival models.

Models considering non-linear relations with the feature set, namely CPH-G and RNN-S, have better accuracy than the other baseline models considered here.
Moreover, only in one instance, sequence length of 4 emails and censoring window of 6 hours, the CPH-G approach has slightly better C-Index than the RNN-S model. 
In all other situations the RNN-S provides higher C-Index than all the baselines.
Thus, in the datasets used here, modeling non-linear relations with features have greater power than considering linear parametric form for the relation or considering mixture distribution of types of recipients.
While performance improvement with increase in length of sequence of past emails is one aspect where neural network benefits prediction, it also adds value for longer duration of censoring window. It is concluded that using sequential information in the neural network has better predictive power than using past information only as features.
    
\subsection{Accuracy in Prediction of Send Time}
\label{sto_section}
In this section we show how RNN-S can be used to compute best times to send emails. Using RNN-S, we calculate the hazard ratio of each individual for the next email, for each send time bin. 
The hazard ratio is related in a monotonically increasing manner to the probability of opening an email within a fixed period of time. For each individual, the different time bins are ranked based on the hazard ratios.
The open rate of emails, if sent during the top $10$ ranked bins for these individuals, is observed to be $19.24\%$. However, the open rate of emails is $16.48\%$, if sent to the same customers during $10$ bins sampled uniformly from all bins. That is a substantial jump of $16.7\%$ in open rate for marketing emails. On a percentage points basis, the $2.76\%$ increment in open rates on $1$ million emails from the sample of $30,000$ customers (data set size for this evaluation) translates to an additional $27,600$ emails being opened, when sent during the appropriate times as predicted by our model.

We now calculate open rates by deciles of send time bins, where the deciles are based on ranks of bins for each individual, as determined by hazard ratios. As shown in Table \ref{sto_results}, there is a marked difference in open rates when emails are sent during the top ranked bins compared to those sent during the lowest ranked bins, which ranges from $19.24\%$ in the top ranked bins to $6.99\%$ in the lowest ranked bins. The bins which are in the lowest deciles are usually in the middle of the night (often these bins cover time from midnight to 5 AM), and depict a marked decrease in open rates. This shows empirically that when emails are sent during inconvenient times, usually they are not opened, even though customers have opportunity to open later. The increasing open rates as the rank of the bins for sending emails increases also indicates that use of our approach to estimate send times can help in increasing open rates.

\begin{table}[!htbp]
\centering
\caption{Open Rates by Deciles of Ranked Bins of Send Time}
\label{sto_results}
\begin{tabular}{|c|c|}
\hline
\multicolumn{1}{|l|}{\textbf{Decile}} & \multicolumn{1}{l|}{\textbf{Open Rates (\%)}} \\ \hline
1                                     & 19.24                                         \\ \hline
2                                     & 18.14                                         \\ \hline
3                                     & 17.48                                         \\ \hline
4                                     & 17.20                                         \\ \hline
5                                     & 16.90                                         \\ \hline
\end{tabular}
\hspace{1cm}
\begin{tabular}{|c|c|}
\hline
\multicolumn{1}{|l|}{\textbf{Decile}} & \multicolumn{1}{l|}{\textbf{Open Rates (\%)}} \\ \hline
6                                     & 16.53                                         \\ \hline
7                                     & 16.21                                         \\ \hline
8                                     & 16.55                                         \\ \hline
9                                     & 15.30                                         \\ \hline
10                                    & 6.99                                          \\ \hline
\end{tabular}
\end{table} 
\section{Discussion and Conclusion}
\label{secConclusion}

Firms want to increase open rates of permission emails. Our focus on a component of emails controlled by firms, namely, their send times, differentiates from prior art. Send times matter since consumers are likely to open marketing emails when they arrive at convenient times and ignore them otherwise. By better predicting times-to-open, our model tunes send times to favorable times-to-open, and can improve likelihood of emails being opened. 
We first predict times-to-open by exploiting the sequential time dependence in individual's historical interactions with emails from a firm. 
We show an extension to survival analysis methods that leverages LSTM networks to incorporate effect of multiple observations on predicting time-to-event. 
The LSTM learns the dependencies in historical interactions to improve prediction of times-to-open. 
In predicting times-to-open, our model outperforms survival analysis methods which ignore the sequential time dependence in email interactions. Moreover, we show that our proposed approach can be successfully used for computing appropriate send times. The current approach gives a simple way of using RNNs for survival analysis. Future work can explore other ways of encoding sequentially time dependent information for survival analysis. %
 
\bibliographystyle{ACM-Reference-Format}
\bibliography{Survival}

\end{document}